\documentclass[10pt,twocolumn,letterpaper]{article}
\pdfoutput=1
\usepackage{iccv}
\usepackage{times}
\usepackage{epsfig}
\usepackage{graphicx}
\usepackage{amsmath}
\usepackage{amssymb}
\usepackage{enumerate}
\usepackage{mathrsfs}
\usepackage{tabularx}
\usepackage{multirow}
\usepackage{float}
\usepackage{comment}
\usepackage{pdfpages}
\usepackage[numbers,sort]{natbib}

\usepackage[pagebackref=true,breaklinks=true,letterpaper=true,colorlinks,bookmarks=false]{hyperref}

 \iccvfinalcopy 

\def\iccvPaperID{593} 

\ificcvfinal\fi
\begin{document}

\title{End-to-End Spatial Transform Face Detection and Recognition}

\author{Liying Chi\\
Zhejiang University\\
{\tt\small charrin0531@gmail.com}
\and
Hongxin Zhang\\
Zhejiang University\\
{\tt\small zhx@cad.zju.edu.cn}
\and
Mingxiu Chen\\
Rokid.inc\\
{\tt\small cmxnono@rokid.com}
}

\maketitle

\begin{abstract}
   Plenty of face detection and recognition methods have been proposed and got delightful results in decades. Common face recognition pipeline consists of: 1) face detection, 2) face alignment, 3) feature extraction, 4) similarity calculation, which are separated and independent from each other. The separated face analyzing stages lead the model redundant calculation and are hard for end-to-end training. In this paper, we proposed a novel end-to-end trainable convolutional network framework for face detection and recognition, in which a geometric transformation matrix was directly learned to align the faces, instead of predicting the facial landmarks. In training stage, our single CNN model is supervised only by face bounding boxes and personal identities, which are publicly available from WIDER FACE \cite{Yang2016} dataset and CASIA-WebFace \cite{Yi2014} dataset. Tested on Face Detection Dataset and Benchmark (FDDB) \cite{Jain2010} dataset and Labeled Face in the Wild (LFW) \cite{Huang2007} dataset, we have achieved 89.24\% recall for face detection task and 98.63\% verification accuracy for face recognition task simultaneously, which are comparable to state-of-the-art results.
\end{abstract}

\section{Introduction}

As the fundamental stage of facial analyzing such as face recognition, age-gender recognition, emotion recognition and face transformation, face detection is an important and classical problem in computer vision and pattern recognition fields.  After the real-time object detection framework proposed by Viola and Jones \cite{Viola2001}, lots of face detection methods have been proposed. The face detection is suffered from many challenges such as illumination, pose, rotation and occlusion. In the past, these challenges were solved by combining different models or using different hand-craft features. In recent years, the Convolutional Neural Network (CNN) presented its powerful ability in computer vision tasks and achieved higher performance\cite{Redmon2016a,Redmon2016}.

The common face recognition methods used faces with known identities to train with classification and took the intermediate layer as feature expression. In the wild, human faces are not always frontal, it is import to extract spatially invariant features from face patches with large pose transformation.Almost all the methods use the face landmarks predictor \cite{Ren2016a,Ren2014,Zhang2016a,Zhang2014,Deng2016,Sun2013}  to locate the position of the face landmarks and then performed the face alignment by fitting the geometric transformation between the predicted facial landmarks and the pre-defined landmarks.

The common pipeline for face recognition consists of: 1) face detection, 2) face alignment, 3) feature extraction, 4) similarity calculation, which are separated and independent from each other. Lots of methods \cite{Liong2013,Lu2015,Stuhlsatz2012} were focusing on how to efficiently extract features from the face patches and make the in-class closer and out-class farther in the feature space. Different loss functions \cite{Zhang2016b,Wen2016} were proposed for this task.

The separated face analyzing stages lead the model redundant calculation and are hard for end-to-end training. Since it has been shown that the jointly learning could boost the performance of individual tasks, such as jointly learning face detection and alignment \cite{Zhang2016a}, many multi-task methods for facial analyzing were proposed \cite{Zhang2016a,Ranjan2016a,Ranjan2016}.

In this paper, a novel end-to-end trainable convolutional network framework for face detection and recognition is proposed. Base on the Faster R-CNN  \cite{Ren2016} framework, this proposed framework could benefit from its strong object detection ability. In the proposed framework, the face landmarks prediction and the alignment stage are replaced by Spatial Transform Network (STN) \cite{Jaderberg2015} , in which a geometric transformation matrix was directly learned to align the faces, instead of predicting the facial landmarks. Compared with facial landmarks prediction network, the STN is smaller and more flexible for almost any feature,  which makes the network end-to-end trainable, and the face detection and recognition tasks could share the common lower features to reduce unnecessary extra feature calculation. This end-to-end network improves the performance and is easy to extend to multi-task problems.

This paper makes the following contributions:

\begin{enumerate}[1]
\item A novel convolutional network framework that is end-to-end trainable for simultaneously face detection and recognition is proposed. In the framework, the STN is used for face alignment, which is trainable and no need to be supervised by labeled facial landmarks.
\item In the proposed framework, the detection part, the recognition part and the STN share common lower features, which makes the model smaller and reduces unnecessary calculations.
\item The single CNN model is supervised only by face bounding boxes and personal identities, which are publicly available from WIDER FACE \cite{Yang2016} dataset and CASIA-WebFace \cite{Yi2014} dataset. Tested on Face Detection Dataset and Benchmark (FDDB) \cite{Jain2010} dataset and Labeled Face in the Wild (LFW) \cite{Huang2007} dataset, the model got 89.24\% recall on the FDDB dataset and 98.63\% accuracy on the CASIA dataset, achieves the state-of-the-art result.
\end{enumerate}

This paper is organized as follows. Related works are introduced in Sec.~\ref{sec2}. The proposed framework is described in detail in Sec.~\ref{sec3}. The experiments results are exhibited in Sec.~\ref{sec4}. Sec.~\ref{sec5} concludes the paper with a brief discussion and feature works.


\section{Related Work} \label{sec2}
  Face Detection develops rapidly in decades. In 2001, Viola and Jones \cite{Viola2001} first proposed a cascade Adaboost framework using the Haar-like features and made the face detection real-time. In recent years, the Convolutional Neural Network has shown its powerful ability in computer vision and pattern recognition. Many CNN-based object detection methods have been proposed \cite{Liu2015,Redmon2016a,Redmon2016,Dai2016,Girshick2016,Ren2016,Girshick2015b} . \cite{Ren2016} improved the region proposal based CNN method and proposed the Faster R-CNN framework, this framework introduced the anchors method and made region proposal a CNN classification problem, which could be trained in the whole net during the training stage. The end-to-end trainable Faster R-CNN network was faster and more powerful, which achieved 73\% mAP in the VOC2007 dataset with VGG-net. \cite{Zheng2016,Jiang2016,Sun2017} used Faster R-CNN framework to solve the face detection problem, and achieved promising results.

Most face recognition methods used aligned faces as the input \cite{Wen2016},  it had been shown that adopting alignment in the test stage could have 1\% recognition accuracy improvement on the LFW \cite{Huang2007} dataset. The usual way for face alignment was predicting facial landmarks from the detected facial patches, such as eyes, nose and mouth. And the geometric transformation between the positions of the predicted facial landmarks and the pre-defined landmarks was applied to the facial patches. The aligned faces with known face identities were then fed into the deep networks and were classified by the last classification layer for training the discriminative feature extractors, the intermediate bottleneck layer was taken as the representation. \cite{Wen2016} proposed a new supervision signal, called Center Loss, for face recognition task, which could simultaneously learn a center for deep features of each class and penalizes the distances between the deep features and their corresponding class centers. Differently, FaceNet \cite{Schroff2015a} used a deep convolutional network to directly optimize the embedding itself, rather than an intermediate bottleneck layer. It used triplets of matching / non-matching face patches generated by a novel online triplet mining method and achieved state-of-the-art face recognition performance using only 128-bytes per face, in their experiments, using the face alignment boosted the accuracy record.

Those methods got high performance on detection and recognition benchmarks, but only focused on the single task. Jointly learning for face detection and alignment first appeared in \cite{Chen2014}, which proved that these two related tasks could benefit from each other. \cite{Zhang2016a} adopted a cascaded structure with three stages of carefully designed deep convolutional networks for predicting face and landmark location simultaneously. \cite{Ranjan2016a} proposed Hyper Face method for simultaneous face detection, facial landmark localization, head pose estimation and gender recognition from a given image, but not included face recognition task. \cite{Ranjan2016} proposed an all-in-one network for learning different facial tasks including face recognition, but the input faces were aligned using Hyper Face, which was not end-to-end trainable from detection to recognition due to the individual face alignment part.

\cite{Jaderberg2015} introduced a new learnable module. The Spatial Transformer, which explicitly allowed the spatial manipulation of data within the network, that could directly learn the best geometric transformation of the input feature maps and transformed features to make them robust for rotation, scale and translation. the STN conducts regression on the transformation parameters directly, and is only supervised by the final recognition objectives.


\section{Methodology} \label{sec3}

\begin{figure*}[!htp]
\begin{center}
\includegraphics[width=1\linewidth]{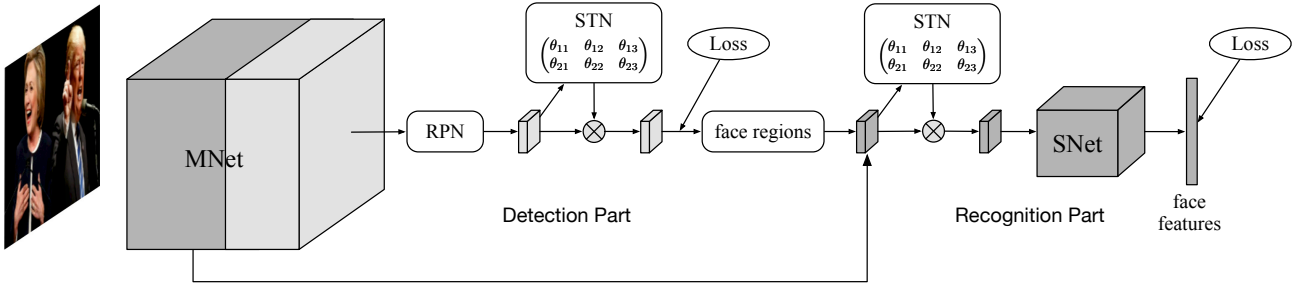}
\end{center}
   \caption{The architecture of the proposed network consists of two parts. The MNet is the main network for extracting the shared features and the detection features, and the SNet is the small network for extracting the facial features from the shared features that used to recognize the person identity. The RPN uses MNet features to generate candidate regions. The STN predicted transformation parameters from the input feature. The SNet uses the transformed features from shared features in detected face regions, and output face features. In the proposed framework, the MNet and SNet could be any CNN network, eg.VGG-16 architecture\cite{Simonyan2014} for MNet and ResNet \cite{He2016} for SNet.}
 \label{fig:architecture}
\end{figure*}

A novel end-to-end trainable convolutional network framework for face detection and recognition is proposed, in which a geometric transformation matrix was directly learned by STN to align the faces, instead of predicting the facial landmarks. Compared with facial landmarks prediction network, the STN is smaller and more flexible for almost any feature, which makes the network end-to-end trainable, and the face detection and the recognition tasks could share the low-level features and reduce unnecessary extra feature calculation. The detection part in proposed architecture is based on the Faster R-CNN framework, which is state-of-the-ate object detector in the world. The recognition part is based on a simplified ResNet \cite{He2016}. In this section, the whole architecture is described into three parts. Sec.~\ref{sec_arch} illustrates the whole architecture. Sec.~\ref{sec_det} describes the detection part. Sec.~\ref{sec_stn} introduces the STN. Sec.~\ref{sec_recog} describes the recognition part.

\subsection{Architecture} \label{sec_arch}

Typically, a face recognition system required the cropped face patches as its input, which are the results of a pre-trained face detector, then the deep CNN processed the cropped faces and obtained the distinguishable features for the recognition task. The common pipeline of face recognition task includes the following stages: 1) face detection, 2) face alignment, 3) feature extraction, 4) similarity calculation. The separated face analyzing stages lead the model redundant calculation and are hard for end-to-end training. In this work, the face detection task and the face recognition task could share the same lower features. By using the STN in face alignment, the model becomes end-to-end trainable because in this case, the gradient can be calculated and backward computation is allowed.

\subsection{Detection Part} \label{sec_det}
Similar to the Faster R-CNN framework \cite{Sun2017}, we use the VGG-16 network \cite{Simonyan2014} as the pre-trained model for face detection, which is pre-trained on ImageNet \cite{Deng2009} for image classification so that we could benefit from its discriminable image representation for different object categories. The region proposal network (RPN), which is a fully convolutional network, uses the convolution results from the VGG network to predict the bounding boxes and the objectless scores for the anchors. The anchors are defined with different scales and ratios for obtaining translation invariance. The RPN outputs a set of proposal regions that have the high probability to be the targets.

In the original Faster R-CNN framework, the ROI-Pooling operator pools the features of the proposal regions which are extracted from the convolutional feature maps into a fixed size. Then the following fully connected layers predict the bounding boxes offset and the scores of the proposal regions. In the proposed method, a spatial transform network (STN) \cite{Chen2016} is applied to implement the region features transformation between the ROI-Pooling layer and the fully connected layers. The STN is used to learn a transformation matrix to make the input features spatially invariant. The STN is flexible that it could be inserted after any convolutional layers without obvious costs in both training and testing. In the detection network, the STN share the features of the proposal regions to regress the transform matrix parameters, and a transform operation uses the predicted parameters to transform the input features. And then the transformed output features will be passed to classification and regression.

\subsection{Spatial Transform Network} \label{sec_stn}

The Convolutional Neural Network shows the surprising performance for feature extraction but is still not effective to be spatially invariant for the input data. For the face analyzing problems, the input faces could be collected in a variety of different conditions such as different scales and rotations. To solve this problem, some methods use a plenty of training data to cover the different conditions, however, almost all the methods used the face landmarks predictor to locate the position of the face landmarks. And performed the face alignment by fitting the geometric transformation between the positions of the predicted facial landmarks and the predefined landmarks. Spatial Transform Network\cite{Jaderberg2015}, which is proposed by DeepMind, allow the spatial manipulation of data within the network. The spatial transform network could learn the invariance of translation, scale, rotation and more generic warping from the feature map itself, without any extra training supervision. Because in STN, the gradient can be calculated and backward computation is allowed, the whole framework becomes end-to-end trainable. The experiments show that the spatial transform network could learn the transform invariance and reduce unnecessary calculation.

For the input feature map $ U \in \mathbb{R}^{H \times W \times C} $, the points could be regard as the transformed result from the aligned feature with $ \theta $. The $ \theta $ could be different  format for different transformation type. In the affine transformation we used in our method, the $\theta$ is a $ 2 \times 3 $ matrix as follow:

\begin{equation}\label{theta}
\theta = 
\left [
\begin{matrix}
\cos\alpha & \textnormal{-}\sin\alpha & t_1
\\ \sin\alpha & \cos\alpha  & t_2
\end{matrix} \right ]
\end{equation}

where $ \alpha $ indicates the rotation angle, and $ t_1, t_2 $ indicate the translate. The pixels in the input feature map is the original points, which are denoted as $(x_i^s, y_i^s)$, and the pixels in the output feature map are the target points that denoted as $ (x_i^t, y_i^t) $. Eq. \eqref{transform} shows the pointwise 2D affine transformation.

\begin{equation}\label{transform}
\begin{pmatrix}
x_i^s
\\ y_i^s
\end{pmatrix}
= \tau_\theta\cdot{l_i^t} =
\left [
\begin{matrix}
\theta_{11} & \theta_{12} & \theta_{13}
\\ \theta_{21} & \theta_{22} & \theta_{23}
\end{matrix} \right ]
\begin{pmatrix}
x_i^t
\\ y_i^t
\\ 1
\end{pmatrix}
\end{equation}

The sampling kernel is applied on each channel of the input feature map $ U $ to obtain the correspond pixel value in the output feature map V. Here we use the bilinear sampling kernel, the process could be written as in Eq. \eqref{bilinear}.

\begin{equation}\label{bilinear}
V_i=\sum_{h=1}^H\sum_{w=1}^WU_{hw}\max(0,1-|x_i^s-w|)\max(0,1-|y_i^s-h|)
\end{equation}

According to \cite{Chen2016}, it is easy to calculate the partial derivatives for $ U,\ x $ and $ y $ to backward the loss and update the parameters. The gradients are defined in Eq. \eqref{derivationU}, Eq. \eqref{derivationX} and Eq. \eqref{derivationX_g}.

\begin{equation}\label{derivationU}
\frac {\partial{V_i}} {\partial{U_{hw}}} = \sum_{h=1}^H\sum_{w=1}^W\max(0,1-|x_i^s-w|)\max(0,1-|y_i^s-h|)
\end{equation}

\begin{equation}\label{derivationX}
\frac {\partial{V_i}} {\partial{x_i^s}} =  \sum_{h=1}^H\sum_{w=1}^WU_{hw}\max(0,1-|y_i^s-h|)g(w,x_i^s)
\end{equation}

\begin{equation}\label{derivationX_g}
g(w,x_i^s) = \left\{
\begin{aligned}
1,  &\ x_i^s \le w < x_i^s + 1 \\
\textnormal{-}1,  &\ x_i^s -1 < w < x_i^s \\
0,  &\ otherwise
\end{aligned}
\right.
\end{equation}

Similar to Eq.\ref{derivationX} for $\frac {\partial{V_i}} {\partial{y_i^s}}$

\subsection{Recognition Part} \label{sec_recog}

After the detection boxes are obtained, the filtered boxes are fed into the following recognition part. Another spatial transform network is added before the recognition part to align the detected faces. The ROI-Pooling operation extracts the features in the detection boxes as before, from the shared feature maps. The STN predicts the transformation parameters and applies the transformation on the region features. The whole network is end-to-end trainable and is supervised only by face bounding boxes and person identities from publicly available datasets.

The architecture of the propose method is shown in Fig.\ref{fig:architecture}. The VGG-16 network feature extractor includes 13 convolutions, and outputs 512 feature maps named $conv5$. The RPN network is a fully convolutional network which outputs the candidate boxes for proposal regions. Then the ROI-Pooling operator extracts the features of the proposal regions from $conv5$ feature maps and resizes it to 7x7, which are the same size of the convolution output in pre-trained VGG model. The spatial transform network for detection includes a convolution layer with $ output=20,\ kernel\ size = 5,\ stride = 1$,  a pooling layer with $ kernel\ size = 2$, and a fully connected layer which is initialized with
$ weight =  0,
bias = \left [ \begin{matrix}1\ 0\ 0\ 0\ 1\ 0 \\ \end{matrix}\right ]^T$.
The following four fully connected layers are set to classify the proposal region and regress the bounding box. The Softmax loss layer and the L2 loss layer are used for supervising the detection training. The spatial transform network for recognition includes a convolution layer with $ output=20,\ kernel\ size = 5,\ stride = 1$,  a pooling layer with $kernel\ size = 2$, and a fully connected layer which is initialized as same.

As the Residual Network (ResNet) showing its generalization ability for lots of machine learning problems, we used the ResNet to extract the distinguishable features of the faces, due to the feature sharing, the ResNet could be simplified. It produces a 512-dimensional output feature vector which is able to capture the stable individual features. Inspired by \cite{Wen2016}, we used the Center Loss function with the Softmax loss for co-operation learning discriminative features for the recognition.


\section{Experiments} \label{sec4}

We designed several experiments to demonstrate the effectiveness of the proposed method. For feature sharing, we share different convolution features to compare the loss decreasing and face recognition accuracy and testing time. Finally, We compared with other methods of face detection recall and face recognition accuracy.

\subsection{Implementation Details}

\begin{figure}
\begin{center}
\includegraphics[width=1\linewidth]{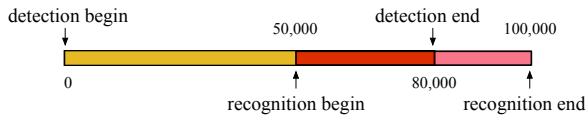}
\end{center}
   \caption{The training stage. The detection network is training before 80,000 iterations, and the recognition network is training from 50,000 iterations.}
\label{fig:training-stage}
\end{figure}

The whole pipeline was written in Python, the CNN was trained on TitanX pascal GPU using the Caffe framework \cite{Jia2014}. we use $base\_lr = 0.1, momentum = 0.9, lr\_policy = step $ with $ gamma = 0.1, stepsize=20,000$. To training more face in one batch, we set $ iter\_size = 32$, the $ max\_iter $ is $ 100,000 $.
The batch size for RPN training was $ 300 $, the positive overlap was set to $ 0.7 $. Due to the different training targets, the two datasets produce different loss\_weight for the backward processing as in Table.\ref{loss_weight}. The detection part was trained for produce accuracy detection results in the first $ 50,000 $ iterations, the detection and recognition part are both trained in the next  $ 30,000 $ iterations, and in the last  $ 20,000 $ iterations, the detection and recognition part were both trained. we fix the detection network parameters and only train the recognition network. The training process is shown in Fig.\ref{fig:training-stage}.

\begin{table}[!htb]
\begin{center}
\begin{tabular}{|c|c|c|c|}
\hline
loss weight & RPN & Detection & Recognition \\
\hline
\centering
WIDER FACE \cite{Yang2016} & 1.0 & 1.0 & 0.0\\
CASIA WebFace \cite{Yi2014} & 0.0 & 0.5 & 1.0 \\

\hline
\end{tabular}
\end{center}
\caption{The loss weight of WIDER Face dataset and CASIA WebFace dataset for different targets.}
\label{loss_weight}
\end{table}

It took around $7$ days to training the network on WIDER FACE dataset and CASIA WebFace dataset. It took around $2$ day to train the first $50,000$ iterations for detection training, and $3$ day for co-training, $2$ day to train the final recognition task. In average the detection took $150ms$ for image size of $1000\times750$. It took about $1ms$ for STN forwards. The feature extraction time per face is shown in Table.\ref{lfw_share_compare}.

\subsection{Dataset}

\subsubsection{Face Detection}

In this work, the WIDER FACE \cite{Yang2016} training set is used for training and the FDDB \cite{Jain2010}dataset is used for testing. WIDER FACE dataset is a publicly available face detection benchmark dataset,  which includes labeled 393,703 faces in 32,203 images. FDDB contains the annotations for 5171 faces in a set of 2845 images taken from the Faces in the Wild data set.


\begin{figure}[htb]
\begin{center}
   \includegraphics[width=0.8\linewidth]{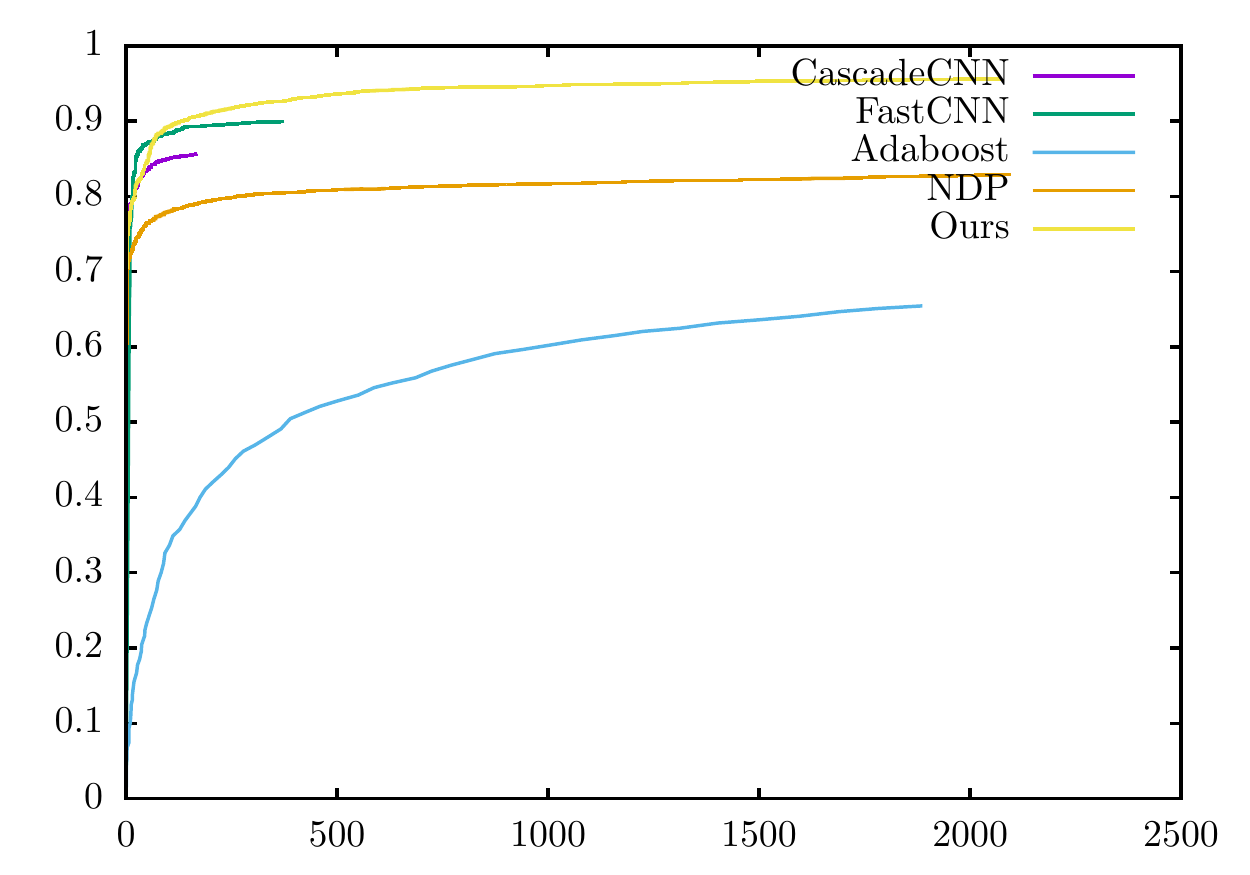}
\end{center}
   \caption{The FDDB results compared with CascadeCNN\cite{Li2015}, FastCNN\cite{Triantafyllidou2016}, Adaboost\cite{Viola2001} and NPD\cite{Liao2016}.}
\label{fig:long}
\label{fig:onecol}
\end{figure}


\begin{figure*}[!htb]
\begin{center}
   \includegraphics[width=1\linewidth]{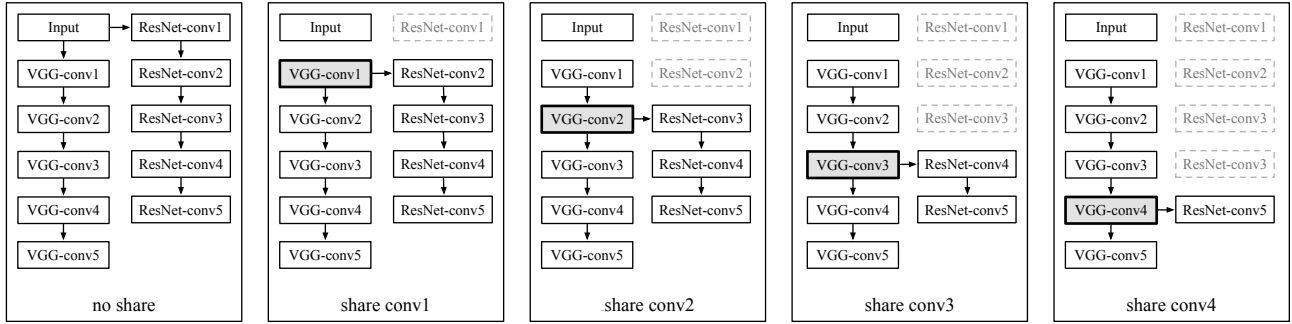}
\end{center}
   \caption{The shared features are illustrated in bold box and the deleted convolution layers in recognition network are illustrated in dotted boxes.}
\label{share_convs}
\end{figure*}

\subsubsection{Face Recognition}

In this work, the model is trained on the CASIA-WebFace dataset which contains 10,575 subjects and 494,414 images collected from the Internet, and tested on LFW \cite{Huang2007} dataset which contains more than 13,000 images of faces collected from the web. Scaling up the count of images in the batch could boost the generalization ability of the model. In the Faster R-CNN framework, the network can use any size of images as the input, but at the same time it loses the ability for processing several images simultaneously. Some methods used the cropped images for the training efficiency, however, to keep the ability for arbitrary input, the images which are randomly selected from the CASIA WebFace dataset are put together. The original image size is $ 250 \times 250 $, by stitching 12 images in 3 rows, for 4 images in each row, the final target training sample size is $ 1000 \times 750 $. Examples of the stitched images are shown in Fig.\ref{casia-add}.

\begin{table}[!htb]
\begin{center}
\begin{tabular}{|l|c|}
\hline
Method & LFW accuracy \\
\hline\hline
DeepID\cite{Sun2014} & 97.45\% \\
VGGFace\cite{Parkhi2015} & 97.27\% \\
\textbf{Ours} & \textbf{98.63\%}\\
\hline
\end{tabular}
\end{center}
\caption{Comparison of single face recognition model on LFW dataset}
\end{table}

\subsection{Share features}
In common face recognition pipeline, the face alignment is applied on the origin input face patches. Benefit from the flexibility of STN which could insert after any convolution layers to align the features, we could share the features for both face detection task and face recognition task. In original detection framework, the first 5 convolution blocks are used for extracting features for face detection task. We designed several training experiments for sharing different features, and tested the results on the FDDB and LFW dataset.

\begin{table}[!htb]
\begin{center}
\begin{tabular}{|c|c|c|}
\hline
                     & LFW results  & test time(ms) \\ \hline
no share &          98.06\% & 8.3 \\ \hline
share conv1 &          98.10\%& 7.8 \\ \hline
share conv2 &          98.23\% & 7.2  \\ \hline
\textbf{share conv3} &  \textbf{98.63\%} & 6.6  \\ \hline
share conv4 &          98.13\%  & 6.3\\ \hline
\end{tabular}
\end{center}
\caption{LFW results of different share convs}
\label{lfw_share_compare}
\end{table}

\begin{figure}[!htb]
\begin{center}
\includegraphics[width=1\linewidth]{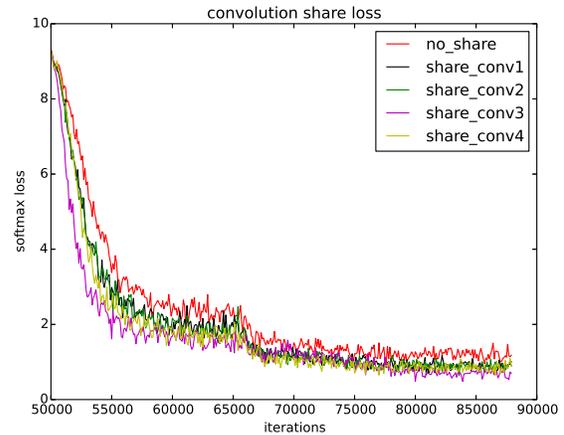}
\end{center}
   \caption{The recognition loss during training for different params.}
\label{fig:recog_loss}
\end{figure}

Fig.\ref{share_convs} illustrates the feature sharing. The shared features include $ conv1, conv2, conv3, conv4 $ of the VGG-16 struct. For each experiment, the recognition networks are simplified by cutting the corresponding convolution layers. The deeper the sharing layers are, the smaller the recognition network will be. Fig.\ref{fig:recog_loss} shows the loss decreasing during the training stage for sharing features. It demonstrates that the deep sharing features help loss convergent faster. Table.\ref{lfw_share_compare} shows the LFW results of the different models. The model gets 98.63\% accuracy on LFW dataset sharing the features of convolution 3, better than using the original image patches. The accuracy decreases when sharing deeper due to the less convolutional layer in the recognition for extracting the distinguishable facial features.

\begin{table*}[!htb]
\begin{center}
\begin{tabular}{|c|c|}
\raisebox{-0.1\height}{\includegraphics[width=0.4\linewidth]{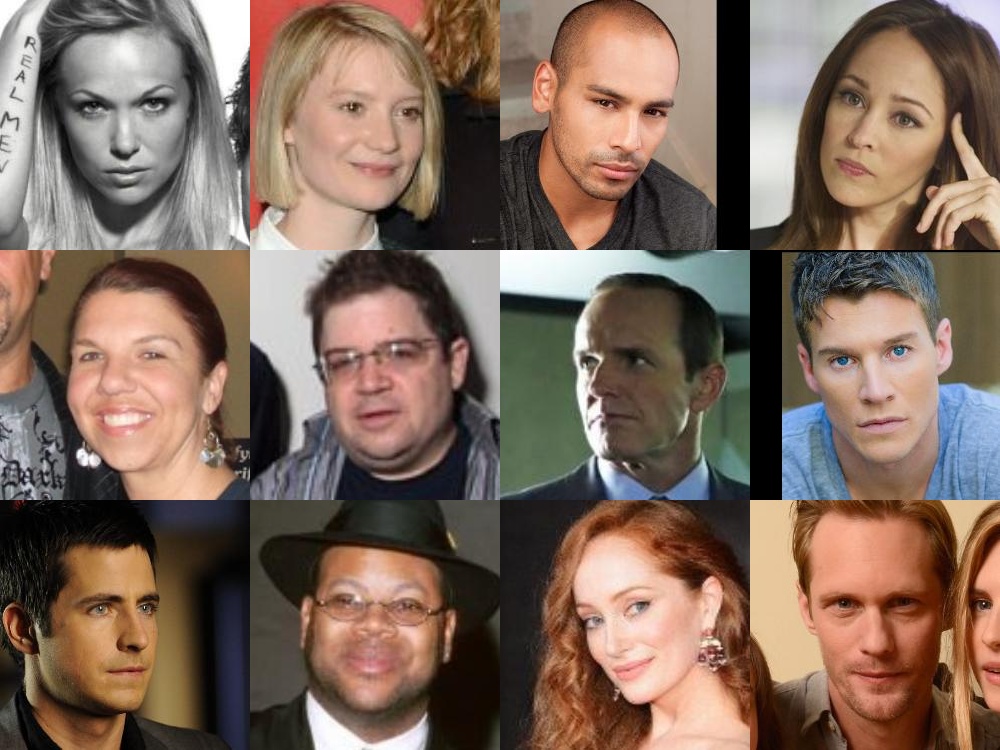}}
\raisebox{-0.1\height}{\includegraphics[width=0.4\linewidth]{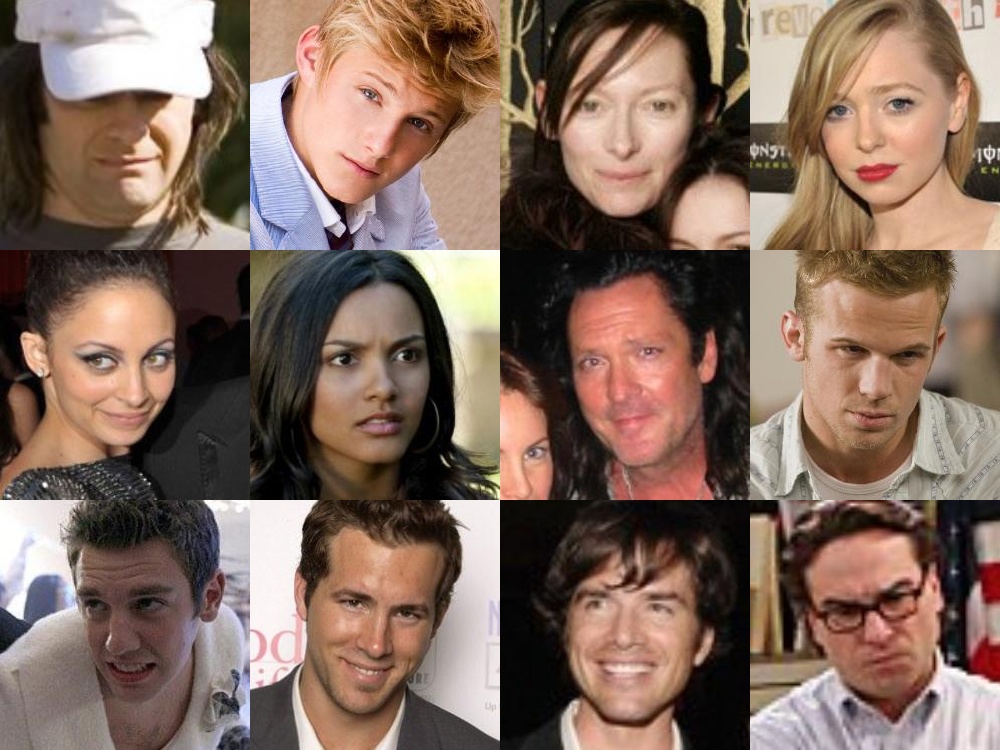}}
\end{tabular}
\end{center}
\caption{The additional dataset for training in the proposed method.}
\label{casia-add}
\end{table*}

\section{Conclusion} \label{sec5}

In this work, a novel end-to-end trainable framework for face detection and face recognition tasks is proposed, in which the spatial transform network is used to align the feature map without additional face alignment stage. The face detection and face recognition network share the low level feature so they could benefit from each other. the experiment demonstrated that the STN could replace the face landmark prediction stage, and sharing common features could make loss convergence faster and more accuracy. Tested on FDDB and LFW dataset, the single model could achieve the state-of-the-art results.

In the future, we are intent to extend this method for other facial targets prediction and make the network size smaller to speed up both training and testing stage.

\begin{table*}[!htb]
\begin{center}
\begin{tabular}{|c|c|c|c|c|}
\raisebox{-0.2\height}{\includegraphics[height=0.18\linewidth]{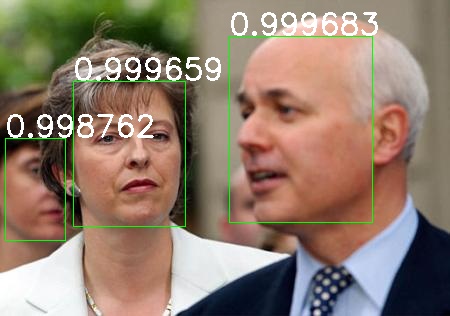}}
\raisebox{-0.2\height}{\includegraphics[height=0.18\linewidth]{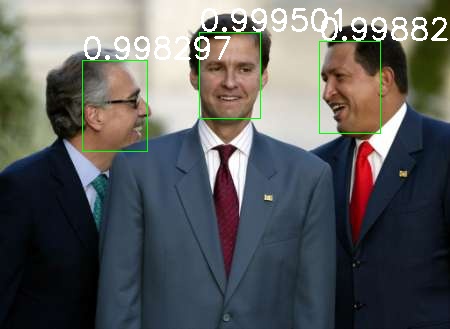}}
\raisebox{-0.2\height}{\includegraphics[height=0.18\linewidth]{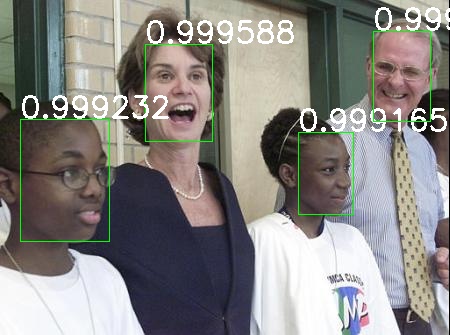}}
\raisebox{-0.2\height}{\includegraphics[height=0.18\linewidth]{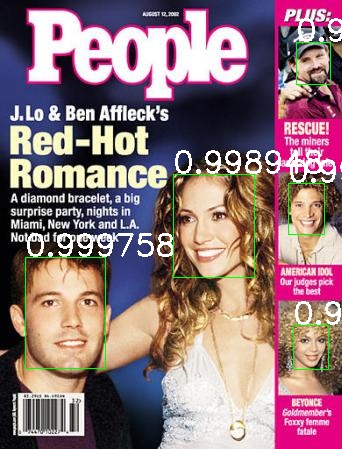}}
\end{tabular}
\end{center}
\caption{Examples of detection results on the FDDB dataset.}
\label{fddb_result}
\end{table*}

\begin{table*}[!htb]
\begin{center}
\begin{tabular}{|c|c|c|c|c|}

\raisebox{-0.1\height}{\includegraphics[width=0.18\linewidth]{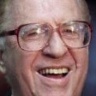}} &
\raisebox{-0.1\height}{\includegraphics[width=0.18\linewidth]{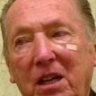}} &
\raisebox{-0.1\height}{\includegraphics[width=0.18\linewidth]{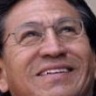}} &
\raisebox{-0.1\height}{\includegraphics[width=0.18\linewidth]{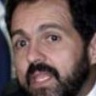}} &
\raisebox{-0.1\height}{\includegraphics[width=0.18\linewidth]{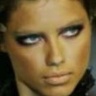}} \\ \hline

\raisebox{-0.1\height}{\includegraphics[width=0.18\linewidth]{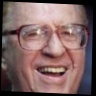}} &
\raisebox{-0.1\height}{\includegraphics[width=0.18\linewidth]{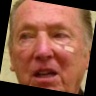}} &
\raisebox{-0.1\height}{\includegraphics[width=0.18\linewidth]{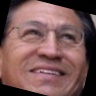}} &
\raisebox{-0.1\height}{\includegraphics[width=0.18\linewidth]{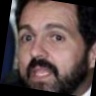}} &
\raisebox{-0.1\height}{\includegraphics[width=0.18\linewidth]{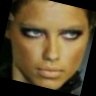}} \\ \hline

\end{tabular}
\end{center}
\caption{Spatial transformation results of CASIA-WebFace dataset. The uppers are the original facial patches. The belows are the transformed results.}
\label{stn_result}
\end{table*}

{\small
\bibliographystyle{ieee}
}

\end{document}